\documentclass[11pt,a4paper]{article}
\usepackage[hyperref]{acl2020}
\usepackage{times}
\usepackage{latexsym}

\usepackage{microtype}
\usepackage{svg}
\usepackage{amsmath}
\usepackage[bottom]{footmisc}
\usepackage{float}
\usepackage[toc,page]{appendix}

\aclfinalcopy 

\title{Continual BERT: Continual Learning for Adaptive Extractive Summarization of COVID-19 Literature}

\author{Jong Won Park \\
  Deerfield Academy \\
  \texttt{jpark21@deerfield.edu}}

\begin{document}
\maketitle
\begin{abstract}
The scientific community continues to publish an overwhelming amount of new research related to COVID-19 on a daily basis, leading to much literature without little to no attention. To aid the community in understanding the rapidly flowing array of COVID-19 literature, we propose a novel BERT architecture that provides a brief yet original summarization of lengthy papers. The model continually learns on new data in online fashion while minimizing catastrophic forgetting, thus fitting to the need of the community. Benchmark and manual examination of its performance show that the model provide a sound summary of new scientific literature\footnote{Our code is available at \url{https://git.io/JJJfO}}.
\end{abstract}

\section{Introduction}
The rapid emergence of the novel coronavirus without much known history has engrossed the international scientific community, resulting in an overwhelming amount of publications and data released on a daily basis. The rate of publications has far exceeded the time-consuming peer-review process, leaving many important information with little to no attention. In an attempt to absorb and utilize the unprecedented amount of COVID-19 scientific literature, prominent journals have opened publications to the public and several platforms have prompted the data science community to aid in the process. One of the notable platform has been COVID-19 Open Research Dataset (CORD-19)\footnote{\url{semanticscholar.org/cord19}} containing thousands of papers published on PubMed and multiple tasks to understand the papers.
\par Recent progress on language processing has made possible the exploration of massive text corpus otherwise infeasible by manual work. Attention-based mechanisms \citep{DBLP:journals/corr/VaswaniSPUJGKP17} and pre-trianed language representations such as BERT \citep{DBLP:journals/corr/abs-1810-04805}, Open GPT-2 \citep{radford2019language}, XLNet \citep{DBLP:journals/corr/abs-1906-08237}, and ELMo \citep{DBLP:journals/corr/abs-1802-05365} have achieved a great success in many language fields, including sentence prediction and text summarization. Many language models are adopting a common practice of pre-training on a huge corpus mined from the web, followed by a fine-tuning process targeted for specific tasks. Following this trend, we focus on utilizing the popular BERT architecture for text summarization task, more specifically extractive summarization, where important sentences are picked from the text verbatim. This task fits the need of the scientific community to rapidly process and extract important information from the inundating number of COVID-19 publications while adhering to their original text.
\par However, as COVID-19 papers are published on a daily basis, many of them with time-sensitive or unseen content, the model also needs to train in an online fashion without experiencing catastrophic forgetting. To this end, we propose \emph{Continual BERT}, a novel BERT architecture built on existing techniques to learn and extract summaries from a continual stream of new tasks while retaining previously learned information. Heavily inspired by \citep{1805.06370}, our architecture utilizes two separate BERT models with layer-wise connections and deploys an alternating training process to minimize catastrophic forgetting. It also stacks a small Transformer encoder on top for extracting summary sentences from text.

\section{Related Work}
\paragraph{Continual Learning}
Recent efforts to train models online, where new data (tasks) flow in a time-sensitive, sequential manner, have increased to fit to the real-world training scenarios. Progressive Neural Networks \citep{DBLP:journals/corr/RusuRDSKKPH16} instantiates new neural networks with layer-wise connections for new tasks, which mitigates catastrophic forgetting but renders inscalable. Progress \& Compress \citep{1805.06370} addresses the scalability by using two separate neural networks with layer-wise connections and online Elastic Weight Consolidation \citep{DBLP:journals/corr/KirkpatrickPRVD16}. Dynamically Expandable Networks \citep{DBLP:journals/corr/abs-1708-01547} takes a different approach by adaptively sizing the neurons in each layer of a network, regularized with group sparse regularization \citep{1607.00485}.
\par Albeit the rapid development on continual learning, few works has focused on incorporating the process onto BERT for language processing. ERNIE 2.0 \citep{DBLP:journals/corr/abs-1907-12412} modifies the pre-training aspect of BERT with a continual learning framework that learns on a broad spectrum of tasks. In contrast, our model modifies the fine-tuning process of BERT for continual learning, which enables leveraging any pre-trained models and focus more on the task-specific fine-tuning process. Alternatively, Adapter-based BERT model was proposed \citep{DBLP:journals/corr/abs-1902-00751} to reduce the training time while retaining old information through learnable adapters. This model also contrasts ours in that it fails to fully leverage the retained previous knowledge in exchange for more efficient training.

\paragraph{Extractive Summarization}
Summarization of text has two categories: extractive and abstractive. The former extracts sentences deemed as a summary, while the latter constructs a unique summary that assimilates the extracted sentences. In the field of extractive summarization, \emph{SummaRuNNer} \citep{DBLP:journals/corr/NallapatiZZ16} uses a sequential model based on Recurrent Neural Network, \emph{LATENT} \citep{zhang-etal-2018-neural-latent} distingushes latent and activated variable sentences and extracts "gold" summaries the latter sentences to improve training, \emph{NeuSum} \citep{DBLP:journals/corr/abs-1807-02305} jointly learns and scores extracted sentences, and \emph{BertSum} \citep{DBLP:journals/corr/abs-1903-10318} stacks Transformer layers on top of BERT for sentence extraction. Although our architecture assimilates BertSum, it differs in that it implements an additional component of continual learning for online tasks.

\begin{table} \label{datasets}
\centering
\begin{tabular}{p{0.3\linewidth}p{0.2\linewidth}p{0.3\linewidth}}
\hline
\textbf{Dataset} & \textbf{Type} & \textbf{Annotated Articles}\\
\hline
NCBI Diesase & Disease & 793 \\
CTD-Pfizer dataset & Drug & 18,410 \\
ScisummNet & General & 1,009 \\
CORD-19 & COVID-19 & 57,037 \\
\hline
\end{tabular}
\caption{Scientific literature dataset with either annotated abstracts or original summaries, including CORD-19 used for our model (as of June 27)}
\end{table}

\begin{figure*}[t]
  \centering
  \includegraphics[width=0.9\textwidth]{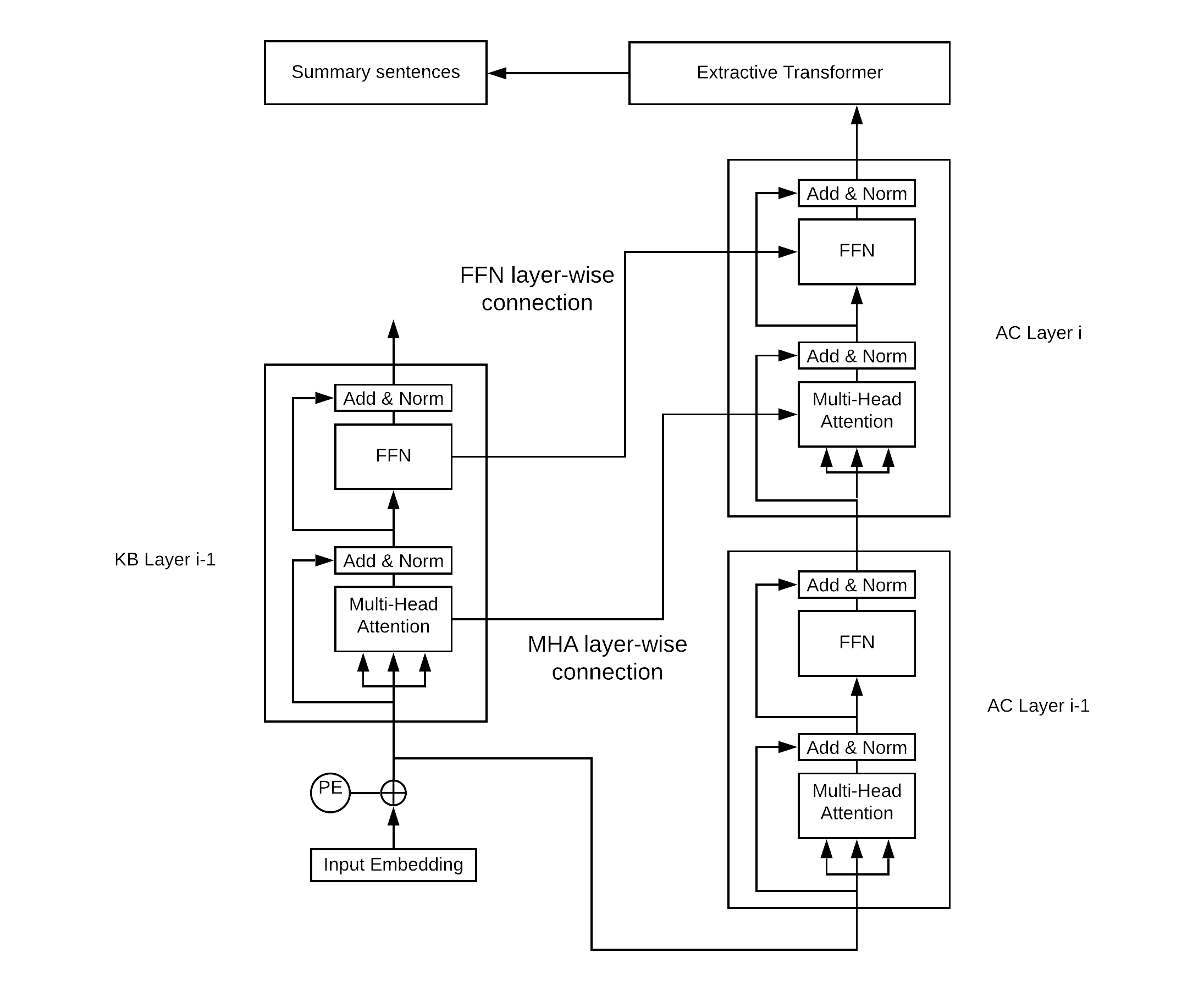}
  \caption{Figure 1. Architecture of \emph{Continual BERT}. Two identical BERT models are initiated, and training on new tasks involves layer-wise connection from Knowledge Base to Active Column. The final output from Active Column is fed into a small, two-layered Transformer stack that outputs the extracted summary sentences. The newly learned parameters are consolidated into the previously learned parameters.}
\end{figure*}

\section{Continual BERT}

\subsection{Architecture}
The base of \emph{Continual BERT} consists of two identical BERT-base models (12-layer, 768-hidden, 12-heads) initialized with the pre-trained \emph{bert-base-uncased} weight. The first BERT-base model is referred to as Active Column (AC), which is trained 'actively' without restrictions. The second BERT-base model is referred to as Knowledge Base (KB), which preserves and contributes previously learned information to AC.
\par The novel feature of \emph{Continual BERT} is the layer-wise connections from KB to AC. More specifically, Multi-Head Attentions (MHA) and Feed-Forward Networks (FFN) of KB layer $i-1$ are wired to that of AC layer $i$ through layer adapters, which are small feed-forward network with trainable parameters. When the model trains on a new task, the new data embedding (calculated with sinusoidal position encoding) is passed onto both KB and AC, and the hidden states in AC are calculated as follows:

\begin{align}
h_{i}=\sigma(W_{i}h_{i-1} + \alpha_{i}\odot{U_{i}\sigma(V_{i}h_{i-1}^{KB} + c_{i})} + b_{i})\nonumber
\end{align}

where $h_{i-1}$ and $h_{i-1}^{KB}$ are the hidden states from layer \emph{i-1} of AC and KB, respectively, $W_i, U_i, V_i$ learnable weights, $b_i, c_i$ biases, $\sigma$ non-linear operation, $\odot$ element-wise multiplication, and $\alpha$ learnable vector. $\alpha$ is initialized from a uniform distribution on the interval $[0,1)$.
\par Thus, for a given task $T$, each AC layer optimize the summation of the parameters learned from the previous task $T-1$ and the parameters connected from KB, which are the compression of all previous tasks $1...T-1$. In essence, the parameters wired from KB add knowledge of all previous tasks and make AC more robust from catastrophic forgetting. The main benefit of these connections is that they enable an aggressive training scheme, which is described in \ref{arch-training}.

\par For extracting summary sentences from literature, we stack a two-layered Transformer encoder on the top of the model. More formally, a two-layered Transformer encoder (768-hidden and 12-heads) computes the importance of each sentences in text with hidden states from the two BERT system.

\subsection{Training} \label{arch-training}
Training is divided into two stages, compress and progress, which are executed in the order listed.
\paragraph{Compress}
During the compress stage, AC trains on a new task $T$ with an exception of layer-wise connections from KB. This provides positive forward transfer from previously tasks $1...T-1$ to benefit the training of the new task data. Unlike the typical procedure of fine-tuning pre-trained BERT models, AC trains aggressively with higher-than-usual hyperparameters (listed in \ref{experiment}, thanks to the retention of all previous tasks in KB. For the training itself, no restrictions are imposed other than the loss function, hence "unrestricted". The model aims to minimize the loss:

\begin{align}
H_p(q) & = -\frac{1}{N}\Sigma_{i=1}^{N}{y_i*\log(p(y_i))}\nonumber\\
        & + (1-y_i)*log(1-p(y_i)\nonumber
\end{align}

where $y$ is the abstract sentence (label).

\paragraph{Progress}

For the consolidation, the model uses online Elastic Weight Consolidation (EWC) \citep{DBLP:journals/corr/KirkpatrickPRVD16} and knowledge distillation. 

After compress, the progress stage begins. Knowledge from AC is distilled into KB in a teacher-student scheme with online Elastic Weight Consolidation (EWC) \citep{DBLP:journals/corr/KirkpatrickPRVD16} to preserve previous information. EWC calculates the Bayesian posterior distribution of parameters using Laplace approximation to calibrate gradient descent towards the overlapping learning region of both the previous and new task. The modified online version of EWC addresses the quadratic cost of the original EWC with a running sum of the diagonal Fisher information matrix and the mean of online Gaussian approximation. The model aims to minimize the online EWC loss:

\begin{align}
L_{EWC} & = \frac{1}{2} \lVert \theta - \theta_{k-1}^{*} \rVert_{\gamma F_{k-1}^{*}}^{2}\nonumber
\end{align}

where $\theta$ is the parameters learned on a new task, $F_{k-1}^{*}$ and $\theta_{k-1}^{*}$ the diagonal Fisher and the mean (of all previous learned parameters) of the online EWC Gaussian approximation, and $\gamma$ hyperparameter to dictate the degradation of previously learned parameters. In addition, the model aims to reduce the knowledge distillation loss:

\begin{align}
L_{KD} & = \mathcal{E}[KL(\pi_{k}(\dot|x)||\pi^{KB}(\dot|x))]\nonumber
\end{align}

where, $x$ is the input, $\pi_{k}(\dot|x)$ and $\pi^{KB}(\dot|x)$ prediction of AC (after learning) and KB, $\mathcal{E}$ expectation over a task dataset. In conclusion, during the progress step, the model minimizes the loss:

\begin{align}
L_{progress} & = L_{EWC} + L_{KD}
\end{align}

\par Since our tasks share the common topic of COVID-19, we preserve the weights of AC through new tasks. By keeping the weights, the model can converge faster with previous parameters over randomly initialized weights. However, if the tasks differ in nature, we recommend re-initializing AC with pre-trained BERT weights for each task to maximize the learning on the new task.

\paragraph{Significance}

The alternating training scheme allows the model be 'online' - continuously learning on new tasks. Online training is critical to the hyperactive ecosystem of literature surrounding COVID-19, where new information can override previously research in a matter of days. Additionally, many new literature rely upon previous data and experiments to reduce the data-collection steps. Thus, any models trained on COVID-19 literature must be versatile to the daily-updating information while still be able to maintain previous knowledge that supplement new findings. By retaining the old knowledge in a compressed form and sharing that with the actively learning column, \emph{Continual BERT} helps to the understanding of the status quo of the SARS-CoV-2 by providing concise and meaningful literature abstracts.

\begin{table*}[t]
\centering
\begin{tabular}{llccc}
\hline
\textbf {Evaluation} & \textbf{Model} & \textbf{ROGUE-1 F} & \textbf{ROGUE-2 F} & \textbf{ROGUE-L} \\
\hline
PubMed (CORD-19) & Continual BERT & \textbf{35.2} & \textbf{15.5} & \textbf{33.8} \\
& BertSum & 30.6 & 11.3 & 28.9 \\
& SummaRuNNer & 20.5 & 8.3 & 9.4 \\
& Adapter-based BERT & 23.6 & 9.2 & 21.8 \\
\hline
ScisummNet & Continual BERT & 31.6 & 12.7 & 30.5 \\
& BertSum & \textbf{33.0} & \textbf{13.4} & \textbf{31.6} \\
& SummaRuNNer & 24.2 & 8.9 & 10.1 \\
& Adapter-based BERT & 25.9 & 10.4 & 24.4 \\
\hline
\end{tabular}
\caption{Evaluation of \emph{Continual BERT} and other similar extractive models on ScisummNet and PubMed (PM) datasets using ROUGE metric. \emph{Continual BERT} achieves significant better evaluation scores on PubMed than ScisummNet. This is attributable to the extensive use of complex and unique medical terms in PubMed compared to the generic scientific terms in ScisummNet.}
\label{table:2}
\end{table*}

\subsection{Dataset}
We use PubMed articles from CORD-19, NCBI Diesase, CTD-Pfizer, and ScisummNet \citep{scisummnet} as our data for the model. We modify the datasets to include only articles with abstracts, resulting in total of 42,000 of 57,037 articles for PubMed, while other dataset retain the same number of articles. The abstracts act as the gold summary for the model evaluation. We train the model to extract up to 20 sentences that are most similar to the abstract. To simulate an online training environment, we divide the articles into tasks of 5,000 articles, ordered by ascending publication time, which results in twelve tasks. The model processes the tasks with older articles first, then at last the task with the most recent articles.

\par We prepare the datasets as a classification task, each classification being a sentence, as proposed by \cite{DBLP:journals/corr/abs-1903-10318}. Each sentence is padded with \emph{[CLS]} at the front to tag it as a classifiable entity and alternating word embedding scheme is used to distinguish adjacent sentences as different sentences. The model learns to classify \emph{[CLS]}, which outputs indice for summary sentences.

\subsection{Experiment} \label{experiment}
The models weights are initialized with Xavier initialization \citep{Glorot2010UnderstandingTD}. For optimization, we use \emph{AdamW} \citep{Loshchilov2019DecoupledWD} with weight decay of $0.01$, $\beta_1 0.9$, $\beta_2 0.999$, and $\epsilon 1e-6$. We use linear learning schedule with learning rate of $5e-4$ and $5\%$ of each task (approximately 300 articles) for warm-up; for knowledge distillation, we use $\tau$ of 2.0 and $\alpha_{ce}$ 0.5; for EWC, we use $\lambda$ of 15 and $\gamma$ 0.99. Other settings include weight decay of $0.01$, batch size of $64$, and $3$ epoch for training AC and KB on each task.

\section{Results}
After learning on twelve tasks ordered in ascending publication time, \emph{Continual BERT} recorded a loss of $0.21$ for compress (consolidation) stage and $2.15$ for progress stage, indicating a difficulty in calibrating to both the previous and new parameters. ROUGE evaluations on PubMed articles and ScisummNet are presented on Table \ref{table:2} and manual evaluation on recent COVID-19 literature is presented on appendix \ref{appendix:manual-eval}. The manual summary evaluation, which is a more realistic and sound technique compared to ROGUE, shows that the model can produce a sound summary of extracted sentences spread throughout the literature. This summary assimilate many sentences provided by the authors, which further supports the model's capability to learn well on new tasks in an online manner.

\section{Discussion}
The online training ability of \emph{Continual BERT} enables adaptive learning on new data flowing in a time-sequential manner, especially fitting to the overwhelming amount of COVID-19 literature published on a daily basis. In contrast to the provided abstracts, extractive summarization of those literature can provide not only a sound, original summary of the article but also indications of where the interesting sentences and ideas lies within the text. This feature can be handy with longer papers, much of COVID-19 literature, as the readers can save significant amount of time while understanding the broad idea of the papers. The scalable architecture of \emph{Continual BERT} also enables continually learning over longer time and in more frequency to digest new research data and information faster.
\par The evaluations listed on Table \ref{table:2} shows that \emph{Continual BERT} performs much better on articles with extensive medical terms (PubMed) compared to the generic scientific articles (ScisummNet). One plausible explanation for this improvement is that PubMed articles from CORD-19 extensively use terms and information related to COVID-19. Furthermore, the generic, coronavirus-related terms appear throughout the decades of PubMed documents. For instance, the medical word "hydroxychloroquine" appears often in many COVID-19 literature. These frequent appearances account for more parameters compressed in KB, thus resulting in the layer-wise connection sharing meaningful information to AC. Hence, the model performs better with more availability of historical knowledge, even if the COVID-19 itself is a new disease.
\par The difficulty that \emph{Continual BERT} experienced during the progress step can be justified with the fact that CORD-19 contains publications dating back to the 20th century, which present radically different information to the more modern publications. This information disparity can be fixed by penalizing more for older publications through time threshold, such as before the coronavirus pandemic.
\par We hope that the model provides a ground for other researchers to explore into the area of summarization for COVID-19 and many other literature. For future explorations, we propose constructing a dynamic version of the model, such as dynamically increasing/decreasing network neurons.

\newpage

\bibliography{acl2020}

\begin{thebibliography}{20}
\expandafter\ifx\csname natexlab\endcsname\relax\def\natexlab#1{#1}\fi

\bibitem[{Devlin et~al.(2018)Devlin, Chang, Lee, and
  Toutanova}]{DBLP:journals/corr/abs-1810-04805}
Jacob Devlin, Ming{-}Wei Chang, Kenton Lee, and Kristina Toutanova. 2018.
\newblock \href {http://arxiv.org/abs/1810.04805} {{BERT:} pre-training of deep
  bidirectional transformers for language understanding}.
\newblock \emph{CoRR}, abs/1810.04805.

\bibitem[{Glorot and Bengio(2010)}]{Glorot2010UnderstandingTD}
Xavier Glorot and Yoshua Bengio. 2010.
\newblock Understanding the difficulty of training deep feedforward neural
  networks.
\newblock In \emph{AISTATS}.

\bibitem[{Houlsby et~al.(2019)Houlsby, Giurgiu, Jastrzebski, Morrone,
  de~Laroussilhe, Gesmundo, Attariyan, and
  Gelly}]{DBLP:journals/corr/abs-1902-00751}
Neil Houlsby, Andrei Giurgiu, Stanislaw Jastrzebski, Bruna Morrone, Quentin
  de~Laroussilhe, Andrea Gesmundo, Mona Attariyan, and Sylvain Gelly. 2019.
\newblock \href {http://arxiv.org/abs/1902.00751} {Parameter-efficient transfer
  learning for {NLP}}.
\newblock \emph{CoRR}, abs/1902.00751.

\bibitem[{Kirkpatrick et~al.(2016)Kirkpatrick, Pascanu, Rabinowitz, Veness,
  Desjardins, Rusu, Milan, Quan, Ramalho, Grabska{-}Barwinska, Hassabis,
  Clopath, Kumaran, and Hadsell}]{DBLP:journals/corr/KirkpatrickPRVD16}
James Kirkpatrick, Razvan Pascanu, Neil~C. Rabinowitz, Joel Veness, Guillaume
  Desjardins, Andrei~A. Rusu, Kieran Milan, John Quan, Tiago Ramalho, Agnieszka
  Grabska{-}Barwinska, Demis Hassabis, Claudia Clopath, Dharshan Kumaran, and
  Raia Hadsell. 2016.
\newblock \href {http://arxiv.org/abs/1612.00796} {Overcoming catastrophic
  forgetting in neural networks}.
\newblock \emph{CoRR}, abs/1612.00796.

\bibitem[{Lan et~al.(2020)Lan, Allan, Malsick, Khandwala, Nyeo, Bathe,
  Griffiths, and Rouskin}]{Lan2020.06.29.178343}
Tammy~C.T. Lan, Matthew~F. Allan, Lauren Malsick, Stuti Khandwala, Sherry~S.Y.
  Nyeo, Mark Bathe, Anthony Griffiths, and Silvi Rouskin. 2020.
\newblock \href {https://doi.org/10.1101/2020.06.29.178343} {Structure of the
  full sars-cov-2 rna genome in infected cells}.
\newblock \emph{bioRxiv}.

\bibitem[{Lee et~al.(2017)Lee, Yoon, Yang, and
  Hwang}]{DBLP:journals/corr/abs-1708-01547}
Jeongtae Lee, Jaehong Yoon, Eunho Yang, and Sung~Ju Hwang. 2017.
\newblock \href {http://arxiv.org/abs/1708.01547} {Lifelong learning with
  dynamically expandable networks}.
\newblock \emph{CoRR}, abs/1708.01547.

\bibitem[{Liu(2019)}]{DBLP:journals/corr/abs-1903-10318}
Yang Liu. 2019.
\newblock \href {http://arxiv.org/abs/1903.10318} {Fine-tune {BERT} for
  extractive summarization}.
\newblock \emph{CoRR}, abs/1903.10318.

\bibitem[{Loshchilov and Hutter(2019)}]{Loshchilov2019DecoupledWD}
Ilya Loshchilov and Frank Hutter. 2019.
\newblock Decoupled weight decay regularization.
\newblock In \emph{ICLR}.

\bibitem[{Nallapati et~al.(2016)Nallapati, Zhai, and
  Zhou}]{DBLP:journals/corr/NallapatiZZ16}
Ramesh Nallapati, Feifei Zhai, and Bowen Zhou. 2016.
\newblock \href {http://arxiv.org/abs/1611.04230} {Summarunner: {A} recurrent
  neural network based sequence model for extractive summarization of
  documents}.
\newblock \emph{CoRR}, abs/1611.04230.

\bibitem[{Peters et~al.(2018)Peters, Neumann, Iyyer, Gardner, Clark, Lee, and
  Zettlemoyer}]{DBLP:journals/corr/abs-1802-05365}
Matthew~E. Peters, Mark Neumann, Mohit Iyyer, Matt Gardner, Christopher Clark,
  Kenton Lee, and Luke Zettlemoyer. 2018.
\newblock \href {http://arxiv.org/abs/1802.05365} {Deep contextualized word
  representations}.
\newblock \emph{CoRR}, abs/1802.05365.

\bibitem[{Radford et~al.(2019)Radford, Wu, Child, Luan, Amodei, and
  Sutskever}]{radford2019language}
Alec Radford, Jeff Wu, Rewon Child, David Luan, Dario Amodei, and Ilya
  Sutskever. 2019.
\newblock Language models are unsupervised multitask learners.

\bibitem[{Rusu et~al.(2016)Rusu, Rabinowitz, Desjardins, Soyer, Kirkpatrick,
  Kavukcuoglu, Pascanu, and Hadsell}]{DBLP:journals/corr/RusuRDSKKPH16}
Andrei~A. Rusu, Neil~C. Rabinowitz, Guillaume Desjardins, Hubert Soyer, James
  Kirkpatrick, Koray Kavukcuoglu, Razvan Pascanu, and Raia Hadsell. 2016.
\newblock \href {http://arxiv.org/abs/1606.04671} {Progressive neural
  networks}.
\newblock \emph{CoRR}, abs/1606.04671.

\bibitem[{Scardapane et~al.(2016)Scardapane, Comminiello, Hussain, and
  Uncini}]{1607.00485}
Simone Scardapane, Danilo Comminiello, Amir Hussain, and Aurelio Uncini. 2016.
\newblock \href {https://doi.org/10.1016/j.neucom.2017.02.029} {Group sparse
  regularization for deep neural networks}.

\bibitem[{Schwarz et~al.(2018)Schwarz, Luketina, Czarnecki, Grabska-Barwinska,
  Teh, Pascanu, and Hadsell}]{1805.06370}
Jonathan Schwarz, Jelena Luketina, Wojciech~M. Czarnecki, Agnieszka
  Grabska-Barwinska, Yee~Whye Teh, Razvan Pascanu, and Raia Hadsell. 2018.
\newblock Progress \& compress: A scalable framework for continual learning.

\bibitem[{Sun et~al.(2019)Sun, Wang, Li, Feng, Tian, Wu, and
  Wang}]{DBLP:journals/corr/abs-1907-12412}
Yu~Sun, Shuohuan Wang, Yu{-}Kun Li, Shikun Feng, Hao Tian, Hua Wu, and Haifeng
  Wang. 2019.
\newblock \href {http://arxiv.org/abs/1907.12412} {{ERNIE} 2.0: {A} continual
  pre-training framework for language understanding}.
\newblock \emph{CoRR}, abs/1907.12412.

\bibitem[{Vaswani et~al.(2017)Vaswani, Shazeer, Parmar, Uszkoreit, Jones,
  Gomez, Kaiser, and Polosukhin}]{DBLP:journals/corr/VaswaniSPUJGKP17}
Ashish Vaswani, Noam Shazeer, Niki Parmar, Jakob Uszkoreit, Llion Jones,
  Aidan~N. Gomez, Lukasz Kaiser, and Illia Polosukhin. 2017.
\newblock \href {http://arxiv.org/abs/1706.03762} {Attention is all you need}.
\newblock \emph{CoRR}, abs/1706.03762.

\bibitem[{Yang et~al.(2019)Yang, Dai, Yang, Carbonell, Salakhutdinov, and
  Le}]{DBLP:journals/corr/abs-1906-08237}
Zhilin Yang, Zihang Dai, Yiming Yang, Jaime~G. Carbonell, Ruslan Salakhutdinov,
  and Quoc~V. Le. 2019.
\newblock \href {http://arxiv.org/abs/1906.08237} {Xlnet: Generalized
  autoregressive pretraining for language understanding}.
\newblock \emph{CoRR}, abs/1906.08237.

\bibitem[{Yasunaga et~al.(2019)Yasunaga, Kasai, Zhang, Fabbri, Li, Friedman,
  and Radev}]{scisummnet}
Michihiro Yasunaga, Jungo Kasai, Rui Zhang, Alexander Fabbri, Irene Li, Dan
  Friedman, and Dragomir Radev. 2019.
\newblock {ScisummNet}: A large annotated corpus and content-impact models for
  scientific paper summarization with citation networks.
\newblock In \emph{Proceedings of AAAI 2019}.

\bibitem[{Zhang et~al.(2018)Zhang, Lapata, Wei, and
  Zhou}]{zhang-etal-2018-neural-latent}
Xingxing Zhang, Mirella Lapata, Furu Wei, and Ming Zhou. 2018.
\newblock \href {https://doi.org/10.18653/v1/D18-1088} {Neural latent
  extractive document summarization}.
\newblock In \emph{Proceedings of the 2018 Conference on Empirical Methods in
  Natural Language Processing}, pages 779--784, Brussels, Belgium. Association
  for Computational Linguistics.

\bibitem[{Zhou et~al.(2018)Zhou, Yang, Wei, Huang, Zhou, and
  Zhao}]{DBLP:journals/corr/abs-1807-02305}
Qingyu Zhou, Nan Yang, Furu Wei, Shaohan Huang, Ming Zhou, and Tiejun Zhao.
  2018.
\newblock \href {http://arxiv.org/abs/1807.02305} {Neural document
  summarization by jointly learning to score and select sentences}.
\newblock \emph{CoRR}, abs/1807.02305.

\end{thebibliography}
\bibliographystyle{acl_natbib}

\newpage

\appendix

\section{Manual evaluation}
\label{appendix:manual-eval}

Manual evaluation on a recent COVID-19 publication published on July 1, 2020. Extracted summary is in lower-case since the pre-trained model is uncased (\emph{bert-base-uncase}).

\vspace*{10px}

\textbf{Title} \\
Structure of the full SARS-CoV-2 RNA genome in infected cells \citep{Lan2020.06.29.178343} (28 pages)

\vspace*{10px}

\textbf{Abstract} \\
SARS-CoV-2 is a betacoronavirus with a single-stranded, positive-sense, 30-kilobase RNA genome responsible for the ongoing COVID-19 pandemic. Currently, there are no antiviral drugs or vaccines with proven efficacy, and development of these treatments are hampered by our limited understanding of the molecular and structural biology of the virus.
Like many other RNA viruses, RNA structures in coronaviruses regulate gene expression and are crucial for viral replication. Although genome and transcriptome data were recently reported, there is to date little experimental data on predicted RNA structures in SARS-CoV-2 and most putative regulatory sequences are uncharacterized. Here we report the secondary structure of the entire SARS-CoV-2 genome in infected cells at single nucleotide resolution using dimethyl sulfate mutational profiling with sequencing (DMS-MaPseq). Our results reveal previously undescribed structures within critical regulatory elements such as the genomic transcription-regulating sequences (TRSs). Contrary to previous studies, our in-cell data show that the structure of the frameshift element, which is a major drug target, is drastically different from prevailing in vitro models. The genomic structure detailed here lays the groundwork for coronavirus RNA biology and will guide the design of SARS-CoV-2 RNA-based therapeutics.

\vspace*{10px}

\textbf{Extracted Summary} \\
sars-cov-2 is an enveloped virus belonging to the genus beta coronavirus, which also includes sars-cov, the virus responsible forthe 2003 sars outbreak, and middle east respiratory syndrome coronavirus (merscov), the virus responsible for the 2012 mers outbreak. despite the devastating effects these viruses have had on public health and the economy, currently no effective antivirals treatment or vaccines exist. there is therefore an urgent need to understand their uniquerna biology and develop new therapeutics against this class of viruses. coronaviruses (covs) have single - stranded and positive - sense genomes that are the largest of all known rna viruses (27 – 32 kb) (masters , 2006). previous studies oncoronavirus structures have focused on several conserved regions that are important forviral replication. for several of these regions, such as the 5’ utr, the 3’ utr , and the frameshift element (fse), structures have been predicted computationally with supportive experimental data from rnase probing and nuclear magnetic resonance (nmr) spectroscopy (plant et al. , 2005; yang and leibowitz , 2015).

\end{document}